\documentclass{applemlr}

\usepackage[utf8]{inputenc}
\usepackage{amsfonts}
\usepackage{amsmath}
\usepackage{amssymb}
\usepackage{nicefrac}
\usepackage{tikz}
\usetikzlibrary{positioning, arrows.meta, shapes.geometric, fit, backgrounds, calc}
\usepackage{pgfplots}
\pgfplotsset{compat=1.18}
\usepackage{wrapfig}
\usepackage{enumitem}
\usepackage{array}
\usepackage{makecell}
\usepackage{colortbl}

\usepackage{caption}                                               
\newcommand{\up}[1]{{\scriptsize\textcolor{teal}{(+#1)}}}

\newcommand{\std}[1]{{\scriptsize$\pm$#1}}

\newcommand{\ours}{DuST}
\newcommand{\appleteaserfigure}{}
\newcommand{\appleappendixtitle}{}

\title{Primal Generation, Dual Judgment:\\ Self-Training from Test-Time Scaling}

\author{
  Yizhu Jiao$^{1,2,\dagger}$ \quad
  Ruixiang Zhang$^{1}$ \quad
  Richard Bai$^{1}$ \quad
  Jiawei Han$^{2}$ \quad
  Ronan Collobert$^{1}$ \quad
  Yizhe Zhang$^{1}$
}

\affiliation{
  Apple$^{1}$
  University of Illinois Urbana-Champaign$^{2}$ \quad
}

\metadata[Correspondence]{\texttt{yizhuj2@illinois.edu}, \texttt{ruixiangz@apple.com}, \texttt{yizhe\_zhang@apple.com}}
\metadata[Note]{$^{\dagger}$Work done at Apple}
\date{\today}

\abstract{\begin{abstract}

Code generation is typically trained in the \emph{primal} space of programs: a model produces a candidate solution and receives sparse execution feedback, often a single pass/fail bit. Test-time scaling enriches the inference procedure by sampling multiple candidates and judging among them, but the comparative information this process reveals is discarded after inference. We argue that this information defines a \emph{dual} judgment space that provides a far richer training signal: the model learns not from an isolated success or failure, but from the relative correctness structure across its own plausible attempts, identifying which succeed, which fail, and what distinguishes them.
We introduce \textbf{\ours{}} (\textbf{Du}al \textbf{S}elf-\textbf{T}raining), a framework for self-training from the dual judgment space. \ours{} samples candidate programs from the model's own distribution, labels them through sandbox execution, retains groups containing both successes and failures, and trains the model to rank candidates by execution correctness using GRPO. The objective is purely discriminative: the model is never directly rewarded for generating correct programs.
Dual self-training improves both judgment and generation. Across five models spanning two families and three scales (4B to 30B), \ours{} consistently improves Best-of-4 test-time scaling on LiveCodeBench. For Qwen3-30B-Thinking on LiveCodeBench v6, judgment quality improves by +6.2 NDCG, single-sample pass@1 improves by +3.1, and Best-of-4 accuracy improves by +4.1. The trained model's single rollout matches the base model's Best-of-4 performance. SFT on the same ranking data improves judgment without improving generation, confirming that on-policy RL is the mechanism that transfers dual-space learning back into primal generation.

\end{abstract}
}

\renewcommand{\mymaketitle}{%
  \tcbset{enhanced,frame hidden}
  \tcbset{colback=titlebg}
  \tcbset{left=0.5cm}
  \tcbset{right=0.5cm}
  \tcbset{top=0.5cm}
  \tcbset{bottom=0.35cm}
  \tcbset{arc=16pt}
  \tcbset{before skip=0pt}
  \tcbset{grow to left by=1.5pt}
  \tcbset{grow to right by=1.5pt}
  \tcbset{overlay={\node[
    anchor=north west,
    at= (frame.north west),
    xshift= 0.5cm,
    yshift= -0.5cm] {\includegraphics[width=0.5cm]{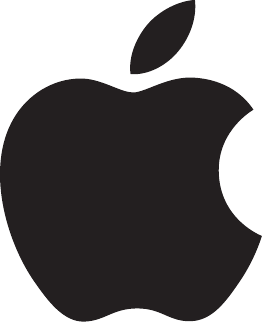}};}}
  \begin{tcolorbox}
    \setlength{\parindent}{0cm}
    \setlength{\parskip}{0.35cm}
    {
      \setlength{\parskip}{0cm}
      \raggedright
      \nohyphens
      {
        \vskip 1.42cm
        \titlelist\par
      }
      \vskip 0.2cm
      \authorlist\par
      \vskip 0.14cm
      \affiliationlist\par
      \contributionlist\par
      \vskip 0.12cm
    }
    \abstractlist\par
    {
      \setlength{\parskip}{0cm}
      \vskip 0.12cm
      \ifdefempty{\metadatalist}{\vspace*{0.2cm}}{\metadatalist\par}
    }
  \end{tcolorbox}
  \tcbset{reset}
  \FloatBarrier
}

\begin{document}

\maketitle

\enlargethispage{0.6cm}
\vspace{0.08cm}
\ifdefined\appleteaserfigure
\begin{center}
\makebox[\textwidth][c]{\includegraphics[width=\dimexpr\textwidth+3pt\relax]{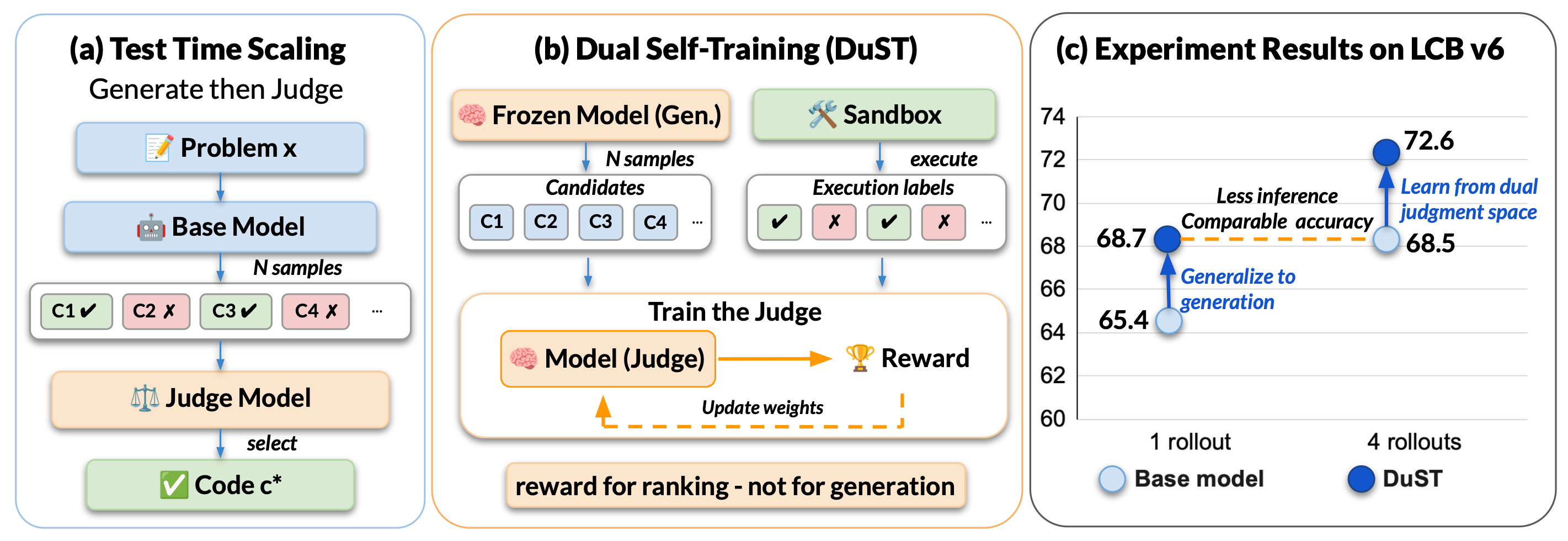}}
\vspace{-0.24cm}
\captionof{figure}{Overview. \textbf{(a)} Conventional test-time scaling samples $N$ candidates and applies a judge, incurring $O(N)$ cost without improving the generator. \textbf{(b)} We propose to self-train from the dual judgment space: the model's own candidates are executed, and the model is trained via GRPO to rank them by correctness---using only a discriminative ranking reward. \textbf{(c)} Using Qwen3-30B-Thinking on LiveCodeBench v6, our proposed method improves the accuracy of test-time scaling via enhancing the judgment quality and surprisingly generalizing to generation. }
\label{fig:intro}
\end{center}
\else
\begin{figure*}[t]
\centering
\includegraphics[width=\textwidth]{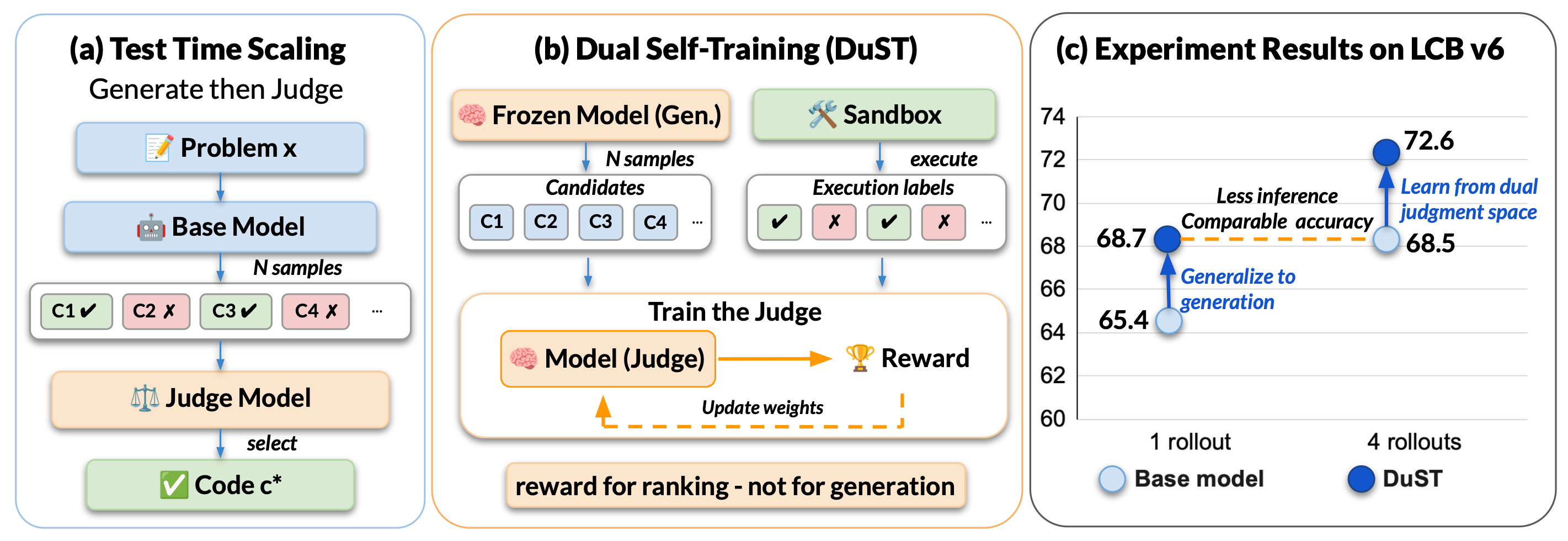}
\label{fig:intro}
\end{figure*}
\fi

\clearpage

\section{Introduction}

\ifdefined\appleteaserfigure
\else
\ifdefined\appleteaserfigure
\begin{center}
\makebox[\textwidth][c]{\includegraphics[width=\dimexpr\textwidth+3pt\relax]{figures/intro_figure.png}}
\vspace{-0.24cm}
\captionof{figure}{Overview. \textbf{(a)} Conventional test-time scaling samples $N$ candidates and applies a judge, incurring $O(N)$ cost without improving the generator. \textbf{(b)} We propose to self-train from the dual judgment space: the model's own candidates are executed, and the model is trained via GRPO to rank them by correctness---using only a discriminative ranking reward. \textbf{(c)} Using Qwen3-30B-Thinking on LiveCodeBench v6, our proposed method improves the accuracy of test-time scaling via enhancing the judgment quality and surprisingly generalizing to generation. }
\label{fig:intro}
\end{center}
\else
\begin{figure*}[t]
\centering
\includegraphics[width=\textwidth]{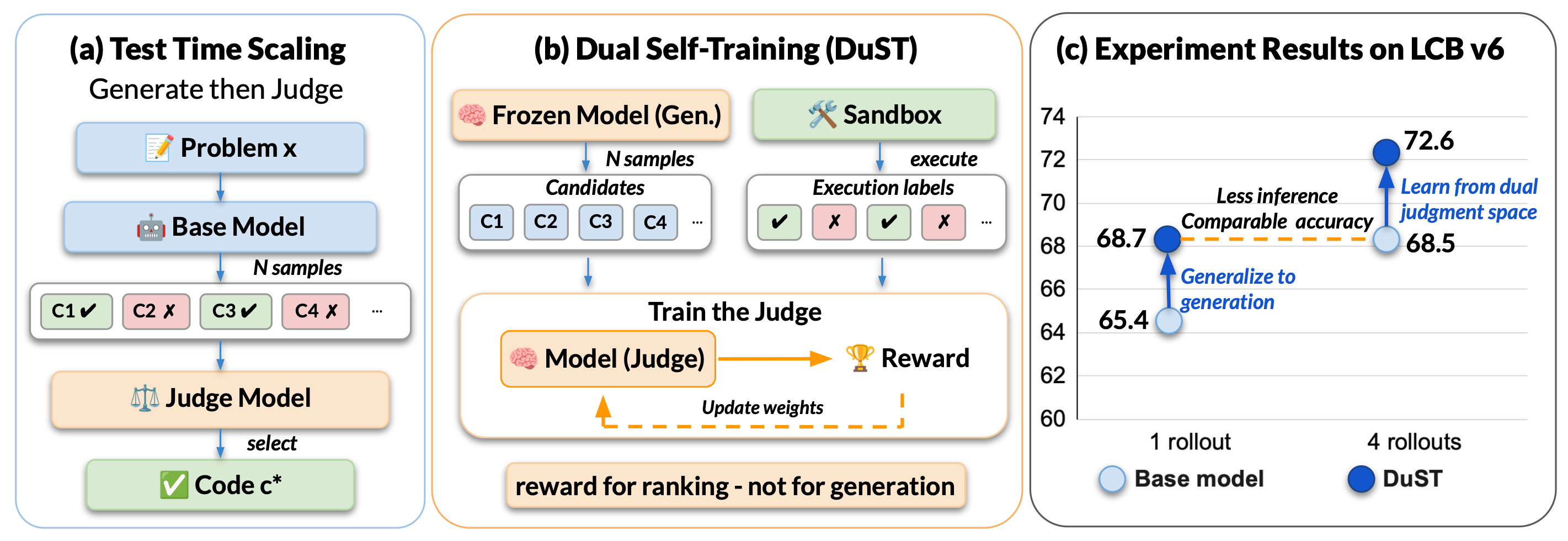}
\label{fig:intro}
\end{figure*}
\fi

\fi

Code generation is often supervised through execution: a model produces candidate code, the code is run against tests, and the resulting feedback indicates whether it is correct. This gives code generation an unusually objective learning signal, but in the standard generation setting, the signal is also highly compressed. A pass/fail outcome tells the model whether one sampled code attempt succeeded, while revealing little about how that attempt compares with other plausible candidates the model could have produced.
Test-time scaling enriches this picture by sampling multiple candidate codes and judging among them \citep{snell2025scaling, yuan2024selfrewarding, zhang2025incentivizing,li2025reasoning, zhou2025evaluating, leerevise, wu2025thought, muennighoff2025s1, wu2025inference}.  In code generation, a common and effective instantiation is the ``generate-then-judge'' recipe: the model samples diverse candidate codes, and a judge ranks them to identify the most promising one \citep{ma2025thinking, yu2025z1, aggarwal2025dars, li2025s, wang2planning}. This paradigm is especially useful in realistic coding settings where hidden tests are unavailable and correctness must be estimated through imperfect proxies. Beyond increasing the chance of sampling correct code, the candidate set exposes a comparative structure over the model's plausible attempts.

Existing generate-then-judge systems typically use this comparative structure only for inference-time ranking, whether through execution feedback, an external judge, or a separately trained reward model \citep{li2022alphacode,zhang2024genrm,toshniwal2026genselect,ma2026agenticverifier}. Once a candidate is selected, however, the signal is discarded. A strong judge can improve the final answer chosen from the current candidate pool, but it does not directly improve the generator that produced those candidates. The model may therefore continue to assign high probability to the same plausible but incorrect code in future rollouts.

This motivates our primal--dual view of test-time scaling. We use the \emph{primal generation space} to refer to the model's ordinary code-generation behavior: sampling candidate codes and receiving execution outcomes on them. We use the \emph{dual judgment space} to refer to the induced ranking problem over those candidates: deciding which plausible attempts are correct, which fail, and what distinguishes them. In the primal space, standard RL training receives only a sparse execution reward for each sampled code attempt, often a single binary pass/fail bit. In the dual space, a set of $N$ candidates can induce up to $O(N^2)$ pairwise comparisons, and every mixed group provides multiple correct-versus-incorrect contrasts. The resulting supervision is therefore much denser than the primal one-bit reward by exposing a richer correctness structure across plausible code attempts rather than an isolated success or failure for each attempt.

We introduce \textbf{\ours{}} (\textbf{Du}al \textbf{S}elf-\textbf{T}raining), a framework for self-training from this dual judgment space. \ours{} samples candidate codes from the model's own distribution, labels them through sandbox execution, retains groups containing both successes and failures, and trains the same model to rank candidates by execution correctness using GRPO \citep{shao2024deepseekmath}. The objective is purely discriminative: the model is never directly rewarded for generating correct code. Instead, all supervision comes from judging correctness contrasts among its own successful and failed attempts. Our central hypothesis is that a model trained to discriminate correct from incorrect code in the dual judgment space can acquire correctness-sensitive features that benefit generation in the primal space.

Our experiments support this hypothesis. On LiveCodeBench v6 \citep{jain2024livecodebench}, Qwen3-30B-Thinking improves by 6.2 NDCG points in judgment quality and by 4.1 points in Best-of-4 test-time scaling accuracy. More importantly, the same training improves single-sample generation by 3.1 pass@1 points, despite using no direct generation reward. This gain generalizes across five base models spanning two model families and three scales (4B to 30B). To isolate the mechanism behind this transfer, we compare GRPO with supervised fine-tuning on the same ranking data. SFT improves judgment but does not improve generation, while GRPO improves both, showing that on-policy reinforcement learning transfers dual-space judgment learning back into primal generation.

\section{Method}
\label{sec:method}

\begin{figure}[t]
\centering
\includegraphics[width=\textwidth]{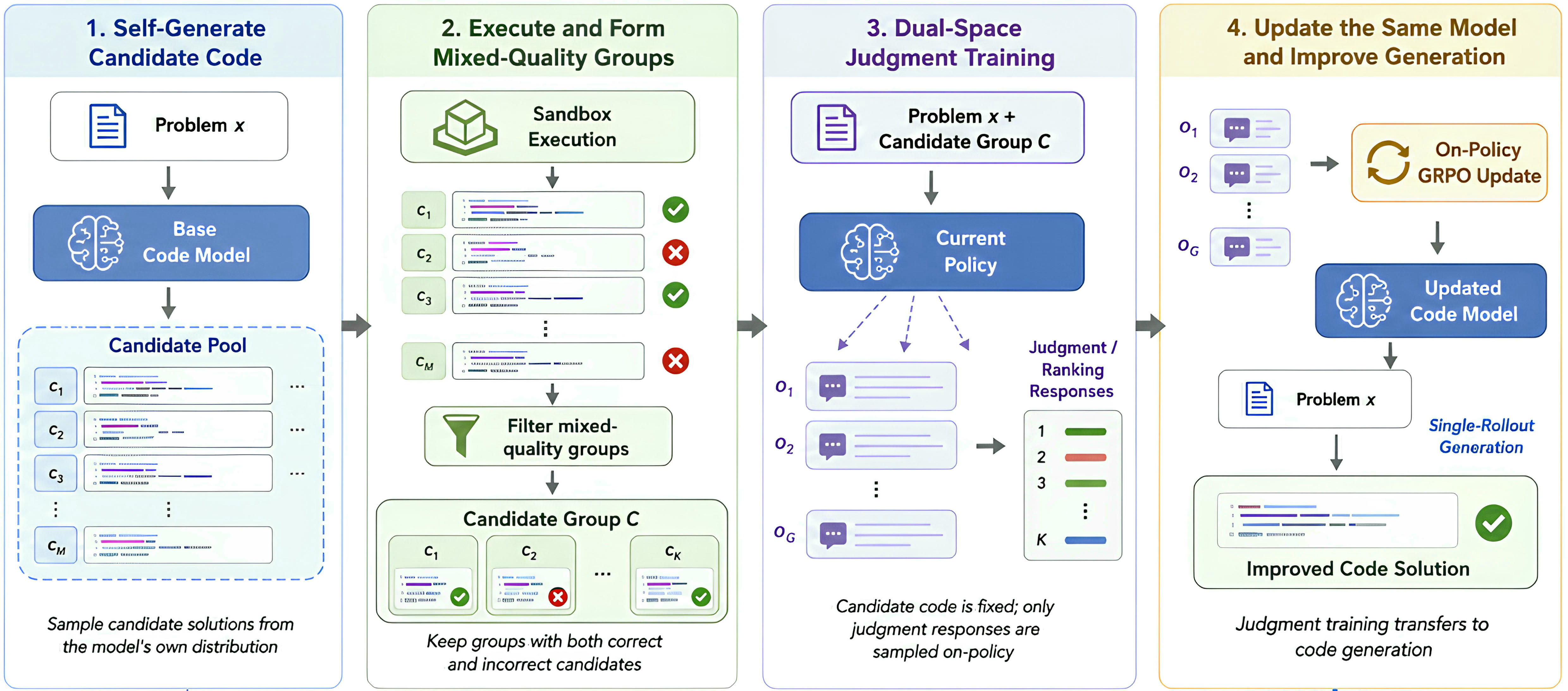}
\caption{Overview of our framework for self-training from dual judgment space. Starting from a base code model, DuST samples candidate code, executes each candidate to obtain binary correctness labels, forms mixed-quality candidate groups, and trains the same model to rank candidates by execution correctness via GRPO. The sampled code is held fixed during training, and on-policy sampling applies only to the model’s judgment responses. Thus, the reward is applied in the dual judgment space, while the updated parameters are the same ones later used for primal code generation.  }
\label{fig:method}
\end{figure}

We now describe \ours{}, a training procedure that converts multi-candidate code generation and judgment into dual-space supervision. Starting from a base code model, \ours{} samples candidate code, executes each candidate to obtain binary correctness labels, forms mixed-quality candidate groups, and trains the same model to rank candidates by execution correctness. The sampled code is held fixed during training, and on-policy sampling applies only to the model's judgment responses. Thus, the reward is applied in the dual judgment space, while the updated parameters are the same ones later used for primal code generation. Figure~\ref{fig:method} summarizes the framework.

\subsection{Test-Time Scaling: From Primal Generation to Dual Judgment}

For a coding problem $x$, a common test-time scaling pipeline proceeds in two stages. First, the model samples a set of candidate code samples
\[
C_N = (c_1,\ldots,c_N),
\qquad
c_i \sim \pi_{\mathrm{base}}(\cdot \mid x).
\]
Second, it applies a judge $J$ to choose a final answer,
\[
c^\star = J(x,C_N).
\]
We refer to this two-stage pipeline as \emph{generate-then-judge}.
The two stages play distinct roles. Sampling $C_N$ expands coverage in the primal generation space, while the judge defines a dual judgment problem over the sampled candidates by deciding which plausible code attempt should be trusted. In standard test-time scaling, however, this comparative correctness information is consumed only at inference time: the judge selects from the current candidate pool, but the signal does not feed back to improve the generator itself.

We instead reuse generate-then-judge as a source of training supervision. Rather than discarding execution-grounded contrasts after selection, \textbf{\ours{}} converts them into dual-space training signals that improve both the model's judgment and its generation, so that a single rollout from the trained model approaches the accuracy of the base model's generate-then-judge pipeline.

\subsection{Constructing Dual-Space Data from Self-Generated Code}
\label{subsec:data}

We construct dual-space training data from the model's own sampled code. For each problem $x$, we sample a large candidate pool from the base model,
\[
\widetilde{C}_M
=
(\tilde{c}_1,\ldots,\tilde{c}_M),
\qquad
\tilde{c}_j \sim \pi_{\mathrm{base}}(\cdot \mid x).
\]
Here $M$ denotes the size of the sampled pool used for data construction, which is distinct from the test-time scaling budget $N$. These candidates are not external demonstrations. They are code samples the model itself considers plausible. The resulting supervision is therefore grounded in what the model can actually generate, rather than on arbitrary positives and negatives.

After execution, each candidate receives a binary correctness label
\[
\tilde{y}_j = V(x,\tilde{c}_j) \in \{0,1\},
\]
where $V(x,c)=1$ indicates that candidate code $c$ passes all tests for problem $x$, and $V(x,c)=0$ otherwise.
We partition the sampled pool into groups of $K$ candidates,
\[
C = (c_1,\ldots,c_K),
\]
with corresponding labels
\[
y = (y_1,\ldots,y_K),
\qquad
y_i = V(x,c_i),
\]
and keep only mixed-quality groups satisfying
\[
0 < \sum_{i=1}^{K} y_i < K.
\]
Each retained group therefore contains at least one correct and one incorrect candidate.

This filtering is useful for two reasons. First, it removes groups that contain no ranking contrast: all-correct groups contain no mistakes to distinguish, while all-incorrect groups contain no execution-confirmed correct code to rank above the failures. Second, it concentrates training on cases where the model can sometimes solve the problem, but cannot reliably do so in every sample. Mixed groups are therefore the basic units of dual-space supervision. They expose objective contrasts between what the model can generate correctly and what it finds plausible but incorrect.

This procedure defines a training set
\[
\mathcal{D}
=
\{(x,C,y)\},
\]
where each instance contains a problem, a self-generated candidate set, and execution-grounded correctness labels. This dataset is constructed once from the base model and reused throughout training. On-policy sampling during GRPO applies only to the model's judgment responses, not to the candidate code samples being judged, making the data pipeline a one-time cost.

\subsection{Dense Execution-Grounded Ranking Rewards}
\label{subsec:reward}

Given a mixed-quality group, the dual-space objective is to rank execution-correct candidates above execution-incorrect candidates. For a training instance $(x,C,y)$, where $C=(c_1,\ldots,c_K)$, the model receives the problem and candidate code samples and outputs a judgment response $o$ containing an optional reasoning trace followed by a final ranking of candidates. Let
\[
\hat{\rho}(o) = (\rho_1,\ldots,\rho_K)
\]
denote the parsed ranking from best to worst, where earlier positions indicate higher predicted correctness.

We use binary execution labels rather than fractional test-pass rates. In code generation, candidate code that fails an edge case is still incorrect, even if it passes most visible tests. Fractional scores can also reflect properties of the test suite rather than the code itself, such as how many tests cover a particular failure mode. Binary correctness gives a sharper supervision signal: the model must distinguish fully correct code from code that contains any latent bug.

We score only the final parsed ranking, not the intermediate reasoning. This keeps the supervision outcome-based: the model may use whatever reasoning strategy it finds useful, but reward is determined solely by whether its final judgment agrees with execution.

We define the set of execution-distinguishable correct-over-incorrect pairs as
\[
\mathcal{P}(y)
=
\{(i,j): y_i=1,\; y_j=0\}.
\]
Since every retained group contains at least one correct and one incorrect candidate, we have $|\mathcal{P}(y)|>0$. Let $\operatorname{pos}_{\hat{\rho}}(i)$ denote the position of candidate $i$ in the predicted ranking. We define the execution-grounded ranking reward as
\[
R_{\mathrm{exec}}(\hat{\rho};y)
=
\frac{1}{|\mathcal{P}(y)|}
\sum_{(i,j)\in \mathcal{P}(y)}
\mathbb{I}
\left[
\operatorname{pos}_{\hat{\rho}}(i)
<
\operatorname{pos}_{\hat{\rho}}(j)
\right].
\]
This reward measures how often correct candidates are ranked above incorrect candidates. Candidates with identical execution labels are not compared, since execution does not specify a preference between them.

To handle malformed judgment responses, we define a response-level reward
\[
R(o;y)
=
\begin{cases}
R_{\mathrm{exec}}(\hat{\rho}(o);y),
& \text{if } o \text{ contains a valid ranking},\\
R_{\mathrm{fmt}},
& \text{otherwise},
\end{cases}
\]
where $R_{\mathrm{fmt}}<0$ is a format penalty.

This listwise formulation provides denser feedback than primal pass/fail supervision. Instead of rewarding each candidate independently, the ranking reward exposes all execution-distinguishable correct-over-incorrect contrasts in the group. With $n_+ = \sum_i y_i$ correct and $n_- = K - n_+$ incorrect candidates per group, the dual reward exposes $n_+n_-$ comparisons, naturally ignoring ties induced by binary labels.

\subsection{On-Policy Dual-Space Self-Training}
\label{subsec:grpo}

The final step is to make the dual judgment signal update the same policy that will later generate code. The candidate code remains fixed, but the judgment responses are sampled from the current policy.

We use Group Relative Policy Optimization (GRPO) \citep{shao2024deepseekmath}. For each training instance $(x,C,y)\sim\mathcal{D}$, we sample a group of $G$ judgment responses from the current policy,
\[
o_1,\ldots,o_G
\sim
\pi_{\theta_{\mathrm{old}}}(\cdot \mid x,C),
\]
score each with the execution-grounded reward $r_i = R(o_i;y)$, and normalize within the group to obtain relative advantages $A_i = (r_i - \bar{r}) / (\operatorname{std}(r_1,\ldots,r_G) + \epsilon)$. The policy is then updated by maximizing the standard clipped GRPO objective with KL regularization against the frozen reference policy $\pi_{\mathrm{ref}}$, using clip ratio $\varepsilon$ and KL coefficient $\beta$.

%

This on-policy step distinguishes \ours{} from supervised imitation of rankings. A supervised model can learn to reproduce correct rankings, but imitation alone need not change the distribution from which code is generated. GRPO instead samples the model's own judgment responses, scores them with execution-grounded rewards, and updates the same parameters used for generation. As a result, correctness-sensitive features learned in the dual judgment task can alter the model's primal generation behavior.

This creates the intended path from dual judgment back to primal generation. To rank candidates correctly, the model must learn features associated with executable correctness, including algorithmic invariants, edge cases, and failure modes. Because these features are learned by the generator itself, they can shift probability mass away from flawed code completions and toward correct code. Thus, any generation improvement is not the result of direct generation supervision, but of the dual judgment signal transferred into the model's generative behavior through on-policy RL.

\section{Experiments}
\label{sec:experiments}

\subsection{Experimental Setup}

\begin{table}[t]
\caption{Test-time scaling performance (accuracy\,\%, mean over 10 repeats) after generating 4 candidates and selecting the top-ranked one, on LiveCodeBench v6 (Feb--May 2025) and v5 (Aug 2024--Feb 2025), stratified by difficulty. Top section lists existing open-source reasoning models sorted by LCB v6 overall score. Bottom section shows our method applied to each base model, with gains over the base shown in parentheses.}
\label{tab:comparison}
\centering
\begingroup
\providecommand{\rowcolor}[1]{}
\renewcommand{\arraystretch}{1.08}
\setlength{\tabcolsep}{4.2pt}
\footnotesize
\newcommand{\tnum}[1]{\makebox[3.0em][r]{#1}}
\newcommand{\tbnum}[1]{\makebox[3.0em][r]{\textbf{#1}}}
\resizebox{0.98\textwidth}{!}{
\begin{tabular}{@{}l*{4}{l}c*{4}{l}@{}}
\toprule
& \multicolumn{4}{c}{\textbf{LiveCodeBench v6}} & & \multicolumn{4}{c}{\textbf{LiveCodeBench v5}} \\
\cmidrule(lr){2-5} \cmidrule(lr){7-10}
\textbf{Model} & \multicolumn{1}{c}{\textbf{Easy}} & \multicolumn{1}{c}{\textbf{Med.}} & \multicolumn{1}{c}{\textbf{Hard}} & \multicolumn{1}{c}{\textbf{All}} & & \multicolumn{1}{c}{\textbf{Easy}} & \multicolumn{1}{c}{\textbf{Med.}} & \multicolumn{1}{c}{\textbf{Hard}} & \multicolumn{1}{c}{\textbf{All}} \\
\midrule
\rowcolor{gray!8}
\multicolumn{10}{@{}l}{\color{gray}\textit{Existing open-source reasoning models}} \\[2pt]
OpenReasoning-Nemotron-32B & \tnum{98.4} & \tnum{73.1} & \tnum{48.0} & \tnum{67.4} & & \tnum{97.6} & \tnum{82.8} & \tnum{53.0} & \tnum{73.2} \\
TinyR1-32B & \tnum{89.5} & \tnum{74.4} & \tnum{46.3} & \tnum{64.9} & & \tnum{88.1} & \tnum{85.6} & \tnum{51.1} & \tnum{71.2} \\
OpenCodeReasoning-1.1-32B & \tnum{97.6} & \tnum{71.2} & \tnum{44.3} & \tnum{64.9} & & \tnum{95.4} & \tnum{83.4} & \tnum{49.4} & \tnum{71.3} \\
OpenReasoning-Nemotron-14B & \tnum{97.6} & \tnum{72.4} & \tnum{39.3} & \tnum{63.0} & & \tnum{95.7} & \tnum{83.4} & \tnum{46.8} & \tnum{70.2} \\
OpenCodeReasoning-1.1-14B & \tnum{96.8} & \tnum{65.4} & \tnum{38.5} & \tnum{60.3} & & \tnum{95.4} & \tnum{76.6} & \tnum{43.3} & \tnum{66.6} \\
OpenReasoning-Nemotron-7B & \tnum{99.2} & \tnum{66.0} & \tnum{34.0} & \tnum{59.0} & & \tnum{97.0} & \tnum{78.0} & \tnum{43.2} & \tnum{67.3} \\
Klear-Reasoner-8B & \tnum{98.4} & \tnum{64.7} & \tnum{32.4} & \tnum{57.6} & & \tnum{96.4} & \tnum{77.3} & \tnum{35.1} & \tnum{63.3} \\
DeepSeek-R1-0528-Qwen3-8B & \tnum{97.6} & \tnum{56.4} & \tnum{29.9} & \tnum{53.8} & & \tnum{96.5} & \tnum{72.6} & \tnum{30.1} & \tnum{59.6} \\
OpenCodeReasoning-1.1-7B & \tnum{94.4} & \tnum{55.8} & \tnum{13.1} & \tnum{45.0} & & \tnum{94.6} & \tnum{60.8} & \tnum{17.2} & \tnum{49.7} \\
Skywork-OR1-7B & \tnum{89.5} & \tnum{41.0} & \tnum{15.2} & \tnum{40.5} & & \tnum{88.1} & \tnum{52.2} & \tnum{20.0} & \tnum{46.8} \\
\midrule
\rowcolor{gray!8}
\multicolumn{10}{@{}l}{\color{gray}\textit{Our method vs.\ base model}} \\[2pt]
Qwen3-30B-Thinking & \tnum{100.0} & \tnum{73.7} & \tnum{47.5} & \tnum{68.7} & & \tnum{98.6} & \tnum{84.4} & \tnum{49.2} & \tnum{72.3} \\
\textbf{+ \ours{}} & \tbnum{100.0} & \tbnum{82.3} & \tbnum{52.5} & \tbnum{72.6} \up{3.9} & & \tbnum{98.6} & \tbnum{87.0} & \tbnum{54.1} & \tbnum{75.2} \up{2.9} \\
\addlinespace
Qwen3-30B-Instruct & \tnum{82.9} & \tnum{49.2} & \tnum{19.0} & \tnum{43.1} & & \tnum{92.4} & \tnum{47.4} & \tnum{19.3} & \tnum{46.0} \\
\textbf{+ \ours{}} & \tbnum{88.2} & \tbnum{49.6} & \tbnum{24.6} & \tbnum{47.1} \up{4.0} & & \tnum{90.2} & \tbnum{53.5} & \tnum{18.7} & \tnum{47.1} \up{1.1} \\
\addlinespace
GPT-OSS-20B & \tnum{98.2} & \tnum{75.2} & \tnum{43.5} & \tnum{65.2} & & \tnum{97.0} & \tnum{79.3} & \tnum{42.9} & \tnum{67.5} \\
\textbf{+ \ours{}} & \tbnum{98.7} & \tbnum{78.5} & \tbnum{47.1} & \tbnum{69.4} \up{4.2} & & \tbnum{97.5} & \tbnum{81.4} & \tbnum{53.1} & \tbnum{72.8} \up{5.3} \\
\addlinespace
Qwen3-4B-Thinking & \tnum{97.6} & \tnum{60.3} & \tnum{29.9} & \tnum{55.0} & & \tnum{95.9} & \tnum{73.3} & \tnum{36.1} & \tnum{62.3} \\
\textbf{+ \ours{}} & \tbnum{98.4} & \tbnum{68.0} & \tbnum{34.0} & \tbnum{59.4} \up{4.4} & & \tbnum{97.9} & \tbnum{77.8} & \tbnum{36.9} & \tbnum{64.6} \up{2.3} \\
\bottomrule
\end{tabular}
}
\endgroup
\end{table}

We apply \ours{} to five models spanning two families (Qwen, GPT-oss), three scales (4B--30B), and two reasoning styles (instruct, thinking): Qwen3-4B-Thinking, Qwen3-14B, Qwen3-30B-Instruct, Qwen3-30B-Thinking, and GPT-oss-20B. Training data is sourced from the seed subset of rSTARcoder \citep{liu2025rstarcoder}, de-duplicated to yield ${\sim}$10K unique competitive programming problems. For each problem, we sample $N{=}64$ candidates from the base model and execute them in a sandboxed environment to obtain binary correctness labels. We partition candidates into groups of $K{=}4$ and retain only mixed-quality groups (at least one correct and one incorrect), yielding ${\sim}$6.9K valid queries and 37K training groups for the primary model; among retained groups, 17.8\% contain one correct candidate, 24.9\% two, and 57.3\% three.

We fine-tune with GRPO \citep{shao2024deepseekmath} using the veRL framework for 3 epochs. Learning rates are $1{\times}10^{-6}$ for Qwen models and $5{\times}10^{-7}$ for GPT-oss-20B; rollout group size is $G{=}8$ with 3 PPO epochs per batch. MoE models train on 32 B200 GPUs with Megatron-based parallelism; dense models on 16 B200 GPUs with FSDP. We evaluate on LiveCodeBench v6 (LCB v6, Feb--May 2025) \citep{jain2024livecodebench} as the primary benchmark and LCB v5 (Aug 2024--Feb 2025) as secondary confirmation, reporting Best-of-4 accuracy and NDCG averaged over 10 repeats. We compare against 10 strong open-source reasoning models (Table~\ref{tab:comparison}). 
More implementation details and prompts are in Appendix~\ref{app:expsetup}.

\subsection{Main Results}

Table~\ref{tab:comparison} compares our method against existing open-source reasoning models on LiveCodeBench v6 and v5. Across all five base models, our method consistently improves test-time scaling accuracy.
The gains are substantial: Qwen3-30B-Thinking improves from 68.7\% to 72.6\% (+3.9) on LCB v6 and from 72.3\% to 75.2\% (+2.9) on LCB v5. Qwen3-4B-Thinking improves from 55.0\% to 59.4\% (+4.4) on LCB v6, Qwen3-30B-Instruct improves from 43.1\% to 47.1\% (+4.0), and GPT-OSS-20B improves from 65.2\% to 69.4\% (+4.2) on LCB v6.
The improvements concentrate on medium and hard problems, where execution-grounded judgment is most valuable: for Qwen3-30B-Thinking on LCB v6, medium accuracy improves from 73.7\% to 82.3\% and hard from 47.5\% to 52.5\%, while easy problems are already saturated near 100\%.
The gains generalize across model families (Qwen and GPT), reasoning styles (thinking and instruct), and scales (4B to 30B), suggesting that execution-grounded ranking reward provides a general-purpose training signal rather than one specific to a particular architecture or scale.

\begin{table}[t]
\caption{Decomposing the effect of dual training on generation (pass@1), judgment (NDCG), and test-time scaling (accuracy). Each row adds one trained component. The gap between rows 2 and 3 isolates the duality effect: the same dual training that improves judgment also improves generation, though generation was never rewarded.}
\label{tab:decomposition}
\centering
\begingroup
\providecommand{\makecell}[1]{\shortstack{#1}}
\providecommand{\rowcolor}[1]{}
\renewcommand{\arraystretch}{1.09}
\setlength{\tabcolsep}{4.4pt}
\footnotesize
\newcommand{\tnum}[1]{\makebox[2.7em][r]{#1}}
\newcommand{\tbnum}[1]{\makebox[2.7em][r]{\textbf{#1}}}
\resizebox{0.98\textwidth}{!}{
\begin{tabular}{@{}l*{6}{l}@{}}
\toprule
& \multicolumn{3}{c}{\textbf{LiveCodeBench v6}} & \multicolumn{3}{c}{\textbf{LiveCodeBench v5}} \\
\cmidrule(lr){2-4} \cmidrule(lr){5-7}
\textbf{Configuration} & \multicolumn{1}{c}{\makecell{Generation\\pass@1}} & \multicolumn{1}{c}{\makecell{Judgment\\NDCG}} & \multicolumn{1}{c}{\makecell{TTS\\Accuracy}} & \multicolumn{1}{c}{\makecell{Generation\\pass@1}} & \multicolumn{1}{c}{\makecell{Judgment\\NDCG}} & \multicolumn{1}{c}{\makecell{TTS\\Accuracy}} \\
\midrule
\rowcolor{gray!8}
\multicolumn{7}{@{}l}{\textit{Qwen3-30B-Thinking}} \\
Baseline & \tnum{65.4} & \tnum{70.1}\std{0.3} & \tnum{68.7} & \tnum{69.2} & \tnum{76.0}\std{0.1} & \tnum{72.3} \\
+ Dual-Trained Judge & \tnum{65.4} & \tnum{73.7}\std{0.3} \up{3.6} & \tnum{70.4} \up{1.7} & \tnum{69.9} & \tnum{76.7}\std{0.1} \up{0.7} & \tnum{74.3} \up{2.0} \\
+ Dual-Trained Gen.\ \& Judge & \tbnum{68.5} \up{3.1} & \tbnum{76.3}\std{0.3} \up{6.2} & \tbnum{72.6} \up{3.9} & \tbnum{71.0} \up{1.8} & \tbnum{78.2}\std{0.2} \up{2.2} & \tbnum{75.2} \up{2.9} \\
\addlinespace[2pt]
\rowcolor{gray!8}
\multicolumn{7}{@{}l}{\textit{Qwen3-30B-Instruct}} \\
Baseline & \tnum{41.1} & \tnum{47.2}\std{0.5} & \tnum{43.1} & \tnum{44.0} & \tnum{50.7} & \tnum{46.0} \\
+ Dual-Trained Judge & \tnum{41.1} & \tnum{48.9}\std{1.0} \up{1.7} & \tnum{45.2} \up{2.1} & \tnum{44.0} & \tnum{51.3} \up{0.6} & \tnum{46.8} \up{0.8} \\
+ Dual-Trained Gen.\ \& Judge & \tbnum{42.5} \up{1.4} & \tbnum{49.4}\std{0.6} \up{2.2} & \tbnum{47.1} \up{4.0} & \tbnum{44.5} \up{0.5} & \tbnum{51.6} \up{0.9} & \tbnum{47.1} \up{1.1} \\
\addlinespace[2pt]
\rowcolor{gray!8}
\multicolumn{7}{@{}l}{\textit{GPT-OSS-20B}} \\
Baseline & \tnum{62.7} & \tnum{70.7}\std{0.2} & \tnum{65.2} & \tnum{67.8} & \tnum{74.9}\std{0.1} & \tnum{67.5} \\
+ Dual-Trained Judge & \tnum{63.5} & \tnum{72.0}\std{0.6} \up{1.3} & \tnum{68.6} \up{3.4} & \tnum{67.7} & \tnum{76.1} \up{1.2} & \tnum{71.7} \up{4.2} \\
+ Dual-Trained Gen.\ \& Judge & \tbnum{65.0} \up{2.3} & \tbnum{72.0}\std{0.3} \up{1.3} & \tbnum{69.4} \up{4.2} & \tbnum{69.8} \up{2.0} & \tbnum{76.9} \up{2.0} & \tbnum{72.8} \up{5.3} \\
\addlinespace[2pt]
\rowcolor{gray!8}
\multicolumn{7}{@{}l}{\textit{Qwen3-4B-Thinking}} \\
Baseline & \tnum{45.8} & \tnum{60.5}\std{0.6} & \tnum{55.0} & \tnum{52.9} & \tnum{67.2}\std{0.3} & \tnum{62.3} \\
+ Dual-Trained Judge & \tnum{47.0} \up{1.2} & \tnum{63.1}\std{0.9} \up{2.6} & \tnum{58.8} \up{3.8} & \tnum{52.4} & \tnum{68.0}\std{0.2} \up{0.8} & \tnum{63.9} \up{1.6} \\
+ Dual-Trained Gen.\ \& Judge & \tbnum{49.1} \up{3.3} & \tbnum{63.0}\std{0.4} \up{2.5} & \tbnum{59.4} \up{4.4} & \tbnum{56.3} \up{3.4} & \tbnum{68.5}\std{0.2} \up{1.3} & \tbnum{64.6} \up{2.3} \\
\bottomrule
\end{tabular}
}
\endgroup
\end{table}

\paragraph{Dual Training Improves Both Judgment and Generation}

To understand where the test-time scaling improvements originate, we decompose the effect of training into generation quality (pass@1), judgment quality (NDCG), and test-time scaling (accuracy) under three configurations: \textit{Baseline} (untrained base model for both generator and ranker), \textit{+\,Dual-Trained Judge} (trained ranker, untrained base generator), and \textit{+\,Dual-Trained Gen.\ \& Judge} (single model for both, trained only with dual judgement training).

\begin{wrapfigure}{r}{0.49\textwidth}
\centering
\captionsetup{font=footnotesize,labelfont=bf}
\includegraphics[width=\linewidth]{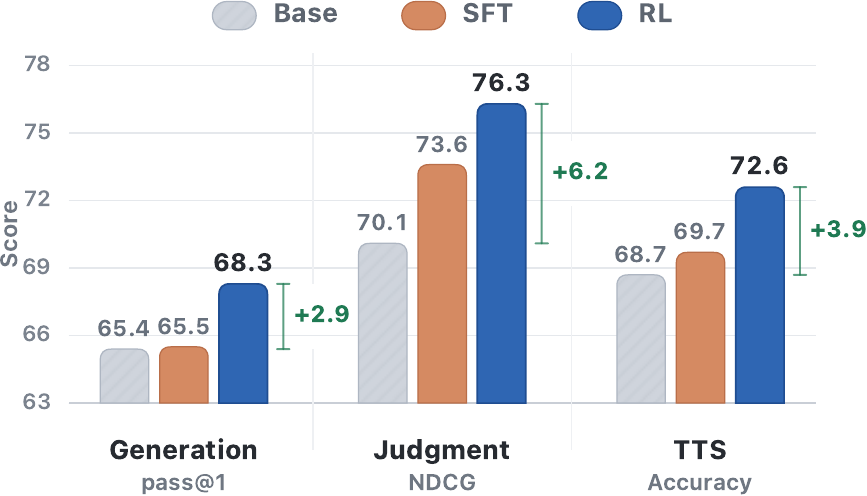}
\caption{Comparison of training modes on LCB v6 (Qwen3-30B-Thinking). SFT improves judgment but not generation; RL improves all three. Here, TTS is short for test time scaling. }
\label{fig:training_modes_bar}
\vspace{-1.0\baselineskip} 
\end{wrapfigure}

Table~\ref{tab:decomposition} reports the decomposition across all four trained models. Two findings stand out. First, dual judgment training consistently improves NDCG across all models (e.g., +6.2 for Qwen3-30B-Thinking, +2.6 for Qwen3-4B-Thinking on LCB v6). Second, the gap between rows 2 and 3 isolates the duality effect: the same dual training that improves judgment also improves primal generation quality despite no direct reward on code correctness. Pass@1 improves for every model (e.g., 65.4\%$\to$68.5\% for Qwen3-30B-Thinking, 62.7\%$\to$65.0\% for GPT-OSS-20B), and the gains from better judgment and better generation compound in the \textit{Dual-Trained Gen.\ \& Judge} configuration.
We hypothesize that this transfer occurs because ranking requires the model to internalize features of executable correctness, like algorithmic invariants, edge cases, implementation bugs. These same features shift the generator toward correct trajectories.

\paragraph{Dual Ranking Training Outperforms Primal Generation Training}

We compare four RL training modes, all using the same base model, data, and reward function (Table~\ref{tab:training_modes}).
\textbf{Off-policy ranking only} (our default) trains in the dual space by ranking pre-generated candidates fixed throughout training.
\textbf{On-policy ranking only} replaces these with candidates sampled from the current policy.
\textbf{On-policy Generation only} removes ranking and trains purely in the primal space with binary execution reward (pass/fail).
\textbf{Iterative (gen+rank)} applies generation and ranking updates at 50 steps iteratively.

The key finding is that dual ranking training outperforms primal generation training across all dimensions. Off-policy ranking achieves the best performance (72.6\% accuracy, 68.3\% pass@1), while generation-only reaches only 71.4\% accuracy and 67.1\% pass@1. This is notable because ranking never directly rewards the model for producing correct code, yet it yields stronger generation than direct execution feedback. The primal binary pass/fail reward is sparser than the dual ranking reward, which exposes the contrastive structure between correct and incorrect solutions through pairwise comparisons per instance.

\paragraph{RL Drives the Dual-to-Primal Transfer}

\begin{wraptable}[13]{r}{0.53\textwidth}                                       
\vspace{-1.6\baselineskip}                                                     
\caption{Four RL training modes on LCB v6 (Qwen3-30B-Thinking), all using      
different rewards. Off-policy ranking (\ours{}) achieves the best performance  
across evaluation dimensions. Here, TTS is short for test time scaling. Gen.    
means generation and Judge. means judgment. }                                  
\label{tab:training_modes}                                                     
\centering                                                                     
\begingroup
\providecommand{\makecell}[1]{\shortstack{#1}}
\renewcommand{\arraystretch}{1.00}
\setlength{\tabcolsep}{10.0pt}
\small                                                                  
\newcommand{\tnum}[1]{\makebox[3.2em][r]{#1}}
\newcommand{\tbnum}[1]{\makebox[3.2em][r]{\textbf{#1}}}                        
\makebox[\linewidth][c]{%
\begin{tabular}{@{}l*{3}{l}@{}}
\toprule                                                                       
\textbf{Training Method} & \multicolumn{1}{c}{\makecell{Gen.\\pass@1}} &
\multicolumn{1}{c}{\makecell{Judge.\\NDCG}} &                                  
\multicolumn{1}{c}{\makecell{TTS\\Accuracy}} \\
\midrule                                                                       
Off-policy Ranking & \tbnum{68.3} & \tbnum{76.3} & \tbnum{72.6} \\
On-policy Ranking & \tnum{67.2} & \tnum{75.3} & \tnum{72.1} \\                 
On-policy Generation & \tnum{67.1} & \tnum{74.6} & \tnum{71.4} \\              
Iterative (gen+rank) & \tnum{66.9} & \tnum{72.4} & \tnum{71.4} \\              
\bottomrule                                                                    
\end{tabular}                                                                  
}%
\endgroup       
\vspace{-2\baselineskip} 
\end{wraptable}            

To isolate whether the transfer from dual judgment to primal generation requires RL's on-policy mechanism or arises from task synergy alone, we compare GRPO against an SFT baseline trained on the same ranking data using Qwen3-235B-Thinking teacher rankings (pairwise accuracy = 1.0), with identical prompt format and data distribution. The SFT model is trained with learning rate $1{\times}10^{-5}$, 3 epochs, and maximum sequence length 65{,}536.
Figure~\ref{fig:training_modes_bar} compares Base, SFT, and RL (GRPO) across all three dimensions. SFT improves judgment (NDCG: 70.1$\to$73.6) and, through better candidate selection from the unchanged base generator, modestly improves accuracy (68.5$\to$70.4). However, SFT leaves generation essentially unchanged (pass@1: 65.4). RL improves all three simultaneously, generation (65.4$\to$68.5), judgment (70.1$\to$76.3), and accuracy (68.5$\to$72.6), because the on-policy feedback loop reshapes the generator as well. We additionally test GRPO after warmup with SFT: RL recovers generation quality (pass@1: 66.9\%, accuracy: 71.8\%), but pure GRPO from base still leads, suggesting SFT initialization constrains RL exploration.
These results confirm that task synergy alone is insufficient. GRPO's closed feedback loop between on-policy sampling, execution feedback, and parameter updates, absent in supervised imitation, is what transfers dual correctness understanding into primal generative behavior.

\begin{figure}[t]
\centering
\begin{tikzpicture}
\definecolor{rewardBlue}{HTML}{3B6DB5}
\definecolor{rewardOrange}{HTML}{C95D3C}
\definecolor{rewardGreen}{HTML}{5B8F5B}
\pgfplotsset{
  rewardAxis/.style={
    width=0.37\textwidth,
    height=0.24\textwidth,
    xmin=10,
    xmax=180,
    xtick={20,60,100,140,180},
    xlabel={Training step},
    axis x line*=bottom,
    axis y line*=left,
    axis line style={draw=black!32,line width=0.6pt},
    tick style={draw=none},
    ymajorgrids=true,
    grid style={draw=black!10,dashed,line width=0.5pt},
    tick label style={font=\scriptsize\sffamily\bfseries,text=black!70},
    xlabel style={font=\scriptsize\sffamily\bfseries,text=black!75,yshift=0.15em},
    title style={font=\scriptsize\sffamily\bfseries,text=black!75,yshift=-0.25em},
    every axis plot/.append style={line width=0.85pt,mark=*,mark size=1.25pt},
    mark options={solid,draw=white,line width=0.35pt},
    clip marker paths=true,
  },
}
\begin{axis}[
  rewardAxis,
  name=gen,
  ymin=63,
  ymax=69,
  ytick={63,65,67,69},
  title={(a) Generation (pass@1)},
]
\addplot+[draw=rewardBlue,mark options={fill=rewardBlue,draw=white,line width=0.35pt}] coordinates {
  (10,64.7) (20,66.3) (30,65.5) (40,65.7) (50,65.2) (60,65.4)
  (70,64.6) (80,66.6) (90,67.5) (100,67.8) (110,67.3) (120,67.8)
  (130,67.3) (140,67.8) (150,68.2) (160,67.2) (170,67.7) (180,67.7)
};
\addplot+[draw=rewardOrange,mark options={fill=rewardOrange,draw=white,line width=0.35pt}] coordinates {
  (10,64.4) (20,65.6) (30,65.1) (40,65.3) (50,64.8) (60,64.9)
  (70,64.4) (80,66.2) (90,66.5) (100,67.0) (110,66.8) (120,67.5)
  (130,66.4) (140,67.2) (150,67.6) (160,66.4) (170,67.1) (180,66.8)
};
\addplot+[draw=rewardGreen,mark options={fill=rewardGreen,draw=white,line width=0.35pt}] coordinates {
  (10,65.8) (20,65.1) (30,64.8) (40,65.2) (50,65.3) (60,65.5)
  (70,65.8) (80,63.6) (90,65.8) (100,65.0) (110,65.2) (120,65.2)
  (130,66.2) (140,65.9) (150,66.4) (160,65.6) (170,65.5) (180,65.1)
};
\end{axis}

\begin{axis}[
  rewardAxis,
  name=judge,
  at={($(gen.east)+(0.03\textwidth,0)$)},
  anchor=west,
  ymin=73,
  ymax=77,
  ytick={73,74,75,76,77},
  title={(b) Judgment (NDCG $\times$ 100)},
]
\addplot+[draw=rewardBlue,mark options={fill=rewardBlue,draw=white,line width=0.35pt}] coordinates {
  (10,74.59) (20,74.65) (30,75.14) (40,75.55) (50,75.75) (60,75.05)
  (70,75.30) (80,75.61) (90,75.01) (100,75.73) (110,74.30) (120,76.12)
  (130,75.78) (140,74.68) (150,75.98) (160,74.84) (170,75.54) (180,75.49)
};
\addplot+[draw=rewardOrange,mark options={fill=rewardOrange,draw=white,line width=0.35pt}] coordinates {
  (10,74.32) (20,74.02) (30,74.18) (40,74.90) (50,75.15) (60,74.33)
  (70,74.58) (80,74.96) (90,74.24) (100,74.77) (110,74.03) (120,75.56)
  (130,75.35) (140,74.20) (150,75.27) (160,74.03) (170,75.29) (180,74.88)
};
\addplot+[draw=rewardGreen,mark options={fill=rewardGreen,draw=white,line width=0.35pt}] coordinates {
  (10,75.47) (20,75.47) (30,74.19) (40,74.36) (50,75.01) (60,75.41)
  (70,74.28) (80,74.07) (90,75.16) (100,75.64) (110,75.17) (120,75.55)
  (130,75.14) (140,75.57) (150,75.81) (160,76.18) (170,75.63) (180,75.90)
};
\end{axis}

\begin{axis}[
  rewardAxis,
  name=tts,
  at={($(judge.east)+(0.03\textwidth,0)$)},
  anchor=west,
  ymin=65,
  ymax=73,
  ytick={65,67,69,71,73},
  title={(c) Test-Time Scaling (accuracy)},
]
\addplot+[draw=rewardBlue,mark options={fill=rewardBlue,draw=white,line width=0.35pt}] coordinates {
  (10,67.6) (20,68.5) (30,68.1) (40,69.8) (50,69.1) (60,68.9)
  (70,68.5) (80,70.4) (90,70.2) (100,71.2) (110,68.9) (120,72.7)
  (130,71.8) (140,69.5) (150,72.0) (160,70.2) (170,71.4) (180,71.6)
};
\addplot+[draw=rewardOrange,mark options={fill=rewardOrange,draw=white,line width=0.35pt}] coordinates {
  (10,66.6) (20,67.8) (30,67.8) (40,69.1) (50,68.6) (60,68.6)
  (70,67.7) (80,70.2) (90,69.3) (100,70.2) (110,68.5) (120,71.9)
  (130,71.1) (140,68.7) (150,71.4) (160,69.5) (170,70.4) (180,71.2)
};
\addplot+[draw=rewardGreen,mark options={fill=rewardGreen,draw=white,line width=0.35pt}] coordinates {
  (10,69.5) (20,69.3) (30,67.8) (40,68.3) (50,69.1) (60,69.5)
  (70,68.7) (80,67.8) (90,70.0) (100,70.2) (110,68.9) (120,69.5)
  (130,69.3) (140,70.0) (150,71.6) (160,72.0) (170,69.5) (180,70.4)
};
\end{axis}

\newcommand{\rewardlegenditem}[2]{%
  \tikz[baseline=-0.45ex]{
    \draw[#1,line width=1.0pt] (0,0) -- (0.36,0);
    \fill[#1] (0.18,0) circle (1.4pt);
  }~#2%
}
\node[anchor=north,font=\scriptsize\sffamily\bfseries,text=black!75,inner sep=0pt]
  at ($(judge.south)+(0,-0.78cm)$)
  {\rewardlegenditem{rewardBlue}{Ranking (PairAcc)}\hspace{1.0em}
   \rewardlegenditem{rewardOrange}{Ranking (NDCG)}\hspace{1.0em}
   \rewardlegenditem{rewardGreen}{Selection (Accuracy)}};
\end{tikzpicture}
\caption{Training dynamics under different judge rewards (Qwen3-30B-Thinking, LiveCodeBench v6). Ranking rewards (PairAcc and NDCG) achieve better generation transfer (pass@1) than Selection reward, which improves judgment but transfers less to generation.}
\label{fig:reward_dynamics}
\end{figure}
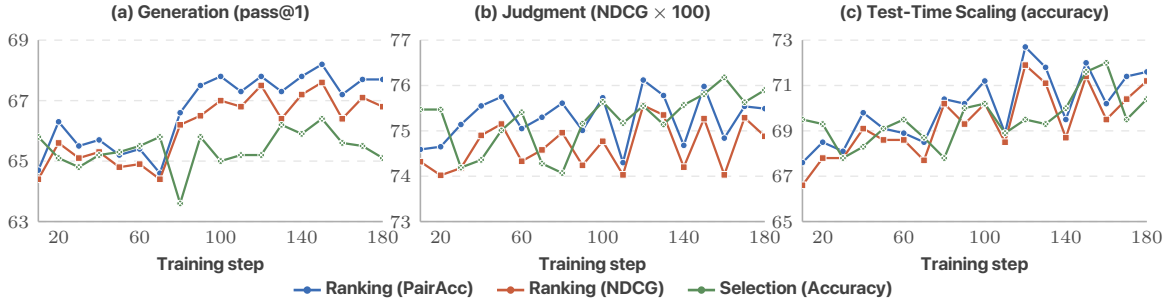

\paragraph{Off-Policy Ranking Outperforms On-Policy Ranking}

Among ranking variants, off-policy outperforms on-policy (72.6\% vs 72.1\%) because the pre-generated pool of $N{=}64$ candidates per problem provides richer contrastive signal, especially for hard problems. On-policy sampling with fewer rollouts is less likely to produce informative mixed-quality groups.
The iterative mode suffers from training instability. Generation updates destabilize training, degrading candidate quality and creating a negative feedback loop that causes early collapse.

\paragraph{Denser Dual Rewards Transfer Better to Generation}

We compare three reward functions for the ranking task (Figure~\ref{fig:reward_dynamics}).
\textbf{Pairwise Accuracy} rewards correct-over-incorrect orderings averaged over all pairs.
\textbf{NDCG} applies position-dependent discounting to the same binary labels.
\textbf{Selection} rewards only if the top-ranked candidate is correct.
Under binary labels, Pairwise Accuracy and NDCG show similar trends, with Pairwise Accuracy achieving a slight edge (72.7\% vs 71.9\% peak accuracy). We adopt it as our default for simplicity.
Selection achieves comparable ranking quality but transfers less to generation: pass@1 stays flat around 65\% to 66\% versus the steady climb to 68\% with ranking rewards.
We attribute this to reward informativeness: ranking requires ordering \emph{all} candidates, exposing the full dual contrastive structure between correct and incorrect code, whereas selection only requires identifying a single best candidate, providing a coarser signal that captures less of this structure.

\section{Related Work}
\label{sec:related_work}

\noindent\textbf{Test-time scaling for code generation.}
Test-time scaling improves model performance by spending extra inference compute on sampling, search, or iterative refinement \citep{snell2025scaling, yuan2024selfrewarding, zhang2025incentivizing,li2025reasoning, zhou2025evaluating, leerevise, wu2025thought, muennighoff2025s1, wu2025inference}. In code generation, this is often instantiated as generate-then-judge: multiple candidate programs are sampled and then ranked or selected using execution feedback, tests, or verifiers, with recent methods further improving diversity, comparison, and repair \citep{ma2025thinking, yu2025z1, aggarwal2025dars, li2025s, wang2planning}. However, these correctness signals are consumed at inference time without enriching the model's primal training signal. \ours{} instead converts them into dual-space supervision, teaching the model to discriminate among its own verified candidates and transferring judgment signals back into primal code generation.

\noindent\textbf{Reinforcement learning for code generation.}
Reinforcement learning has been widely used to improve code generation by optimizing execution-grounded signals that are not directly captured by next-token prediction, such as compilation success, unit-test pass rates, functional correctness, or feedback from learned critics and verifiers \citep{le2022coderl, shojaeeexecution}. Recent work extends this idea by using execution feedback for multi-turn refinement, constructing denser process-level rewards, or training critics and reward models to guide program repair and candidate selection \citep{zeng2025acecoder, gehring2025rlef, ye2025process, li2025codeprm, xie2025teaching}. These methods typically optimize models to generate better code, revise code using feedback, or provide verification signals. In contrast, \ours{} trains exclusively in the dual judgment space: execution labels correctness contrasts among the model's own candidates, and GRPO optimizes a discriminative ranking objective---never directly rewarding primal generation. We find that this dual-space training transfers back into primal generation, improving single-rollout code quality alongside test-time scaling accuracy.

\noindent\textbf{Synergy of generation and judgment.}
Generation and judgment have become deeply intertwined in preference-based language model training. RLHF and DPO-style methods use human or AI preferences to optimize generators \citep{ouyang2022training, rafailov2023direct, yuan2023rrhf, meng2024simpo}, while self-improvement methods use model-generated critiques, revisions, or rewards as feedback \citep{bai2022constitutional, yuan2024selfrewarding, wuself}. These works show that judgment can guide generation, but their feedback is typically external, subjective, or model-generated. Conversely, recent judge-training methods use generation to improve judgment, asking models to produce rationales, critiques, plans, or reasoning traces before assigning scores \citep{saha2501learning, song2024mind, ye2025learning, yu2025self}. In contrast, \ours{} asks whether judgment learning itself can become a route to better generation. We study this question in code generation, where judgment can be grounded in objective execution outcomes. This distinguishes \ours{} from preference optimization, which aligns generation to external preferences; from judge-training work, which primarily improves evaluation; and from standard code RL, which directly optimizes program success. Our focus is whether discriminative judgment learning can transfer back to primal code generation.

\section{Conclusion}

We introduced \ours{}, a framework that self-trains code models in the dual judgment space exposed by test-time scaling. Rather than learning from isolated pass/fail feedback in the primal space, \ours{} trains the model on-policy to discriminate correct from incorrect candidates---never directly rewarding generation. The core premise is that generation and judgment are dual: a model that learns to discriminate correct from incorrect code should also produce better code, since understanding what makes a solution succeed informs how to generate one. Our experiments confirm this across five models: \ours{} consistently improves judgment, generation, and test-time scaling accuracy, with a single rollout from the trained model matching the base model's Best-of-4 performance. SFT on the same data improves judgment but not generation, confirming that on-policy RL is the mechanism that transfers dual-space learning back into primal generation.

These findings suggest that the primal and dual spaces of code generation are more tightly coupled than standard training assumes. The contrastive correctness structure exposed at test time---which trajectories succeed, which fail, and what distinguishes them---is a rich and largely untapped training signal. Exploiting this duality more broadly, through iterative self-improvement, harder problem curricula, or richer dual objectives, is a promising direction for future work.

{
\small
\bibliographystyle{plainnat}
\bibliography{references}
}

\appendix

\newpage
\clearpage

\ifdefined\appleappendixtitle
\noindent{\Huge\sffamily\bfseries Appendix\par}
\vspace{0.65cm}
\fi

\section{Limitations}
\label{appendix:discussions}

We discuss the limitations of this work from four perspectives: the execution-grounded scope of our supervision, the dependence on informative candidate sets, the coverage of our empirical evaluation, and the incomplete understanding of judgment-to-generation transfer.

\subsection{Execution-Grounded Scope}
\label{sec:limitations_execution}

\textbf{\ours{} relies on execution-grounded correctness labels, which makes the framework especially suitable for code generation but less directly applicable to open-ended generation tasks.}
A key advantage of code generation is that candidate solutions can be executed in a sandbox and labeled with objective pass/fail outcomes. This enables us to construct dense pairwise correctness contrasts without relying on human preferences or model-generated judgments. However, many generation domains do not admit such clean verification. For tasks such as open-ended writing, dialogue, or complex reasoning without executable checks, correctness is often subjective, underspecified, or only partially observable. Extending the dual self-training idea beyond code therefore requires reliable task-specific verifiers or carefully designed proxy signals. Our results should thus be interpreted as evidence for execution-grounded domains rather than a general claim that discriminative self-training always transfers to generation.

\subsection{Dependence on Mixed-Quality Candidate Sets}
\label{sec:limitations_candidate_set}

\textbf{The effectiveness of \ours{} depends on the model producing candidate groups that contain both successes and failures.}
Our training signal comes from correctness contrasts among the model's own sampled candidates. This makes the method naturally self-adaptive, but it also introduces a dependence on the quality and diversity of the candidate pool. If the base model is too weak, most candidate groups may contain only failures, providing little positive signal for ranking. If the model is already very strong on a task, candidate groups may contain mostly successes, leaving few informative negative examples. The dual judgment space becomes less informative. The method is therefore most effective in a regime where the model can occasionally solve a problem but still produces plausible incorrect attempts, allowing execution feedback to reveal meaningful correctness boundaries.

\subsection{Imperfect Verification}
\label{sec:limitations_verification}

\textbf{Execution feedback is objective but not always a perfect measure of semantic correctness.}
Sandbox execution provides a stronger supervision signal than free-form preference judgments, but it is still bounded by the quality and coverage of the available tests. A candidate that passes the provided tests may still fail hidden edge cases, rely on brittle assumptions, or exploit accidental properties of the test suite. Conversely, a candidate may fail due to minor formatting or environment issues despite containing largely correct algorithmic logic. Since \ours{} uses execution outcomes to construct ranking labels, such imperfections can introduce noise into the dual judgment signal. Although our evaluation uses held-out benchmark tests to measure final performance, stronger verification protocols would further reduce the risk that the model learns benchmark-specific or test-specific shortcuts.

\subsection{Mechanism and Training Stability}
\label{sec:limitations_mechanism}

\textbf{Our results show that GRPO transfers discriminative judgment learning to generation, but the precise mechanism remains only partially understood.}
The contrast between GRPO and SFT suggests that on-policy reinforcement learning is critical: SFT improves judgment but does not yield the same generative gains, while GRPO improves both. However, this does not fully explain how discriminative ranking updates reshape the model's generation distribution. The transfer may involve changes in internal correctness features, probability mass shifts away from flawed reasoning trajectories, or improved exploration induced by on-policy optimization. Moreover, RL training can be sensitive to candidate sampling strategy, reward design, group construction, and optimization hyperparameters. A deeper mechanistic analysis of when and why dual-space training improves primal generation remains an important direction for future work.

\section{Experiment Setup}\label{app:expsetup}

\noindent\textbf{Data synthesis.}
We use the seed subset of the rSTARcoder dataset \citep{liu2025rstarcoder}, de-duplicated to yield ${\sim}$10K unique competitive programming problems.
For each prompt, we sample $N{=}64$ candidate solutions from the base model using vLLM with temperature 1.0, top-$p$ 1.0, top-$k$ 0, and 128K maximum sequence length.
Each candidate is executed in a sandboxed environment against the problem's test suite to obtain a binary correctness label (pass/fail).
We apply minimal syntactic filtering to remove empty responses and failed code extractions, meaning there is absolutely no correctness signal used during generation.
We then partition the 64 candidates per problem into groups of $K{=}4$ and retain only \emph{mixed-quality} groups containing at least one correct and one incorrect candidate, discarding trivially solved or unsolved problems.
This yields approximately 6.9K valid queries and 37K training groups for the primary model (Qwen3-30B-Thinking). Among the retained groups, 17.8\% contain one correct candidate, 24.9\% contain two, and 57.3\% contain three.

\begin{table}[ht]
\caption{Full GRPO training configuration when using Qwen3-30B-A3B-Thinking as the base model. The Mixture-of-Experts model requires expert parallelism.}
\vspace{2mm}
\centering
\small
\begin{tabular}{ll}
\toprule
\textbf{Parameter} & \textbf{Value} \\
\midrule
Rollouts per prompt ($K$) & 8 \\
Global batch size & 64 prompts $\times$ 8 rollouts = 512 \\
PPO mini-batch & 16 \\
PPO micro-batch per GPU & 2 \\
PPO epochs per batch & 1 \\
Learning rate & $1 \times 10^{-6}$ \\
KL regularization ($\beta$) & 0 (disabled) \\
Clip ratio & [0.2, 0.28] \\
Total data epochs & 3 \\
Advantage normalization & Group std (GRPO) \\
Max sequence length & 32768 (prompt) + 32768 (response) \\
Checkpoint frequency & Every 10 steps \\
\midrule
\multicolumn{2}{l}{\textit{Infrastructure}} \\
Actor parallelism & TP=4, PP=1, EP=8, CP=1 \\
vLLM rollout & TP=4, GPU util.\ 0.7 \\
Hardware & 2 nodes $\times$ 8 B200 GPUs = 32 GPUs \\
\bottomrule
\end{tabular}
\end{table}

\noindent\textbf{Training.}

We fine-tune with GRPO \citep{shao2024deepseekmath} using the VERL framework \footnote{\url{https://github.com/verl-project/verl}}.
The rollout group size is $G{=}8$, with 3 PPO epochs per batch, clip ratio 0.2, and KL coefficient $\beta{=}0.001$.
We train for q epochs over the data.
For Qwen models, we use a learning rate $1{\times}10^{-6}$ and a global batch size of 64 to 128 prompts. For GPT-OSS-20B we use $5{\times}10^{-7}$ and batch size 32.
MoE models (Qwen3-30B variants) use Megatron-based parallelism (EP=8) on 32 B200 GPUs across 4 nodes. The dense Qwen3-14B trains on 16 B200 GPUs across 2 nodes with FSDP, and GPT-OSS-20B trains on 16 B200 GPUs with FSDP2.
On-policy rollouts are served by vLLM with temperature 1.0, top-$p$ 1.0, and 0.7 GPU memory utilization.

\noindent\textbf{Evaluation.}
Our primary benchmark is LiveCodeBench v6 (LCB v6, Feb to May 2025) \citep{jain2024livecodebench}, comprising 131 problems stratified by difficulty (31 easy, 39 medium, 61 hard).
We report LCB v5 (Aug 2024 to Feb 2025, following the version split adopted by rSTARcoder \citep{liu2025rstarcoder}) as a secondary confirmation.
For each evaluation run, the generator samples four candidate codes per problem using temperature 0.6, top-$p$ 0.95, top-$k$ 20, and maximum generation length 65{,}536 tokens. The ranker orders the four candidates, and the top-ranked solution is executed against test cases.
We report the mean performance over 10 repeats.
Our primary metrics are pass@1 for code generation and NDCG for code ranking.
For direct generation, pass@1 measures the probability that a single sample is correct.
For test-time scaling, the generator produces four candidate codes, the ranker selects the top-ranked one, and accuracy is computed on the selected solution.
We choose 10 strong open-source reasoning models as baselines, as listed in Table \ref{tab:comparison}, including OpenReasoning-Nemotron-32B \citep{ahmad2025opencodereasoningiisimpletesttime}, TinyR1-32B \citep{tinyr1proj, si2025efficientswitchablesafetycontrol}, OpenCodeReasoning-1.1-32B \citep{ahmad2025opencodereasoning}, OpenReasoning-Nemotron-14B \citep{ahmad2025opencodereasoningiisimpletesttime}, OpenCodeReasoning-1.1-14B \citep{ahmad2025opencodereasoning}, OpenReasoning-Nemotron-7B \citep{ahmad2025opencodereasoningiisimpletesttime}, Klear-Reasoner-8B \citep{DBLP:journals/corr/abs-2508-07629}, DeepSeek-R1-0528-Qwen3-8B \citep{DBLP:journals/corr/abs-2501-12948}, OpenCodeReasoning-1.1-7B \citep{ahmad2025opencodereasoning}, Skywork-OR1-7B \citep{DBLP:journals/corr/abs-2505-22312}. We evaluate those baselines using its optimal parameters using the consistent prompts with our proposed method. For each baseline, we run it for 10 times for the average evaluation scores.

\section{Prompt Design}\label{app:prompt}

\subsection{Code Ranking Prompt}

The ranking prompt presents the model with a coding problem and $K$ candidate solutions, and asks the model to rank them from most to least likely to pass all tests. The model is free to produce an optional reasoning trace before outputting the final ranking; reward is computed solely on the final parsed ranking.

The prompt template is as follows:

\begin{quote}
\ttfamily\small
You will be given a question (problem specification) and \{$K$\} candidate solutions. You will analyze these candidates and RANK them from most likely to least likely to pass all tests. You will NOT return anything except for the ranking.

\medskip
Question: \{problem\_description\}

\medskip
\{candidate\_solutions\}

\medskip
Analyze each candidate solution, analyze all the strengths and weaknesses, and rank them from most effective to least effective. Output ONLY the ranking as comma-separated numbers from best to worst (e.g., ``2, 4, 1, 3'').

\medskip
Ranking (best to worst):
\end{quote}

Each candidate is formatted as:

\begin{quote}
\ttfamily\small
--- CANDIDATE \{i\} ---\\
```python\\
\{code\}\\
```
\end{quote}

The ranking output is parsed as a comma-separated sequence of candidate indices. If the output does not contain a valid ranking (e.g., due to format errors or repeated indices), the response receives the format penalty $R_{\mathrm{fmt}} = -1$.

\subsection{Code Generation Prompt}

During data synthesis and evaluation, the model generates code solutions using the following prompt template. For stdin-type problems:

\begin{quote}
\ttfamily\small
You will be given a question (problem specification) and will generate a correct Python program that matches the specification and passes all tests. You will NOT return anything except for the program.

\medskip
Question: \{question\}

\medskip
Read the inputs from stdin, solve the problem, and write the answer to stdout (do not directly test on the sample inputs). Enclose your code within delimiters as follows. Ensure that when the Python program runs, it reads the inputs, runs the algorithm, and writes output to STDOUT.\\
```python\\
~~\# YOUR CODE HERE\\
```
\end{quote}

For function-type problems with starter code:

\begin{quote}
\ttfamily\small
You will be given a question (problem specification) and will generate a correct Python program that matches the specification and passes all tests. You will NOT return anything except for the program.

\medskip
Question: \{question\}

\medskip
You will use the following starter code to write the solution to the problem and enclose your code within delimiters.\\
```python\\
\{starter\_code\}\\
```
\end{quote}

For thinking models (e.g., Qwen3-30B-Thinking), the instruction ``You will NOT return anything except for the program'' is omitted, allowing the model to reason before the code block.

\end{document}